\pdfoutput=1
\documentclass[11pt]{article}

\usepackage{naacl2021}

\usepackage{times}
\usepackage{latexsym}
\usepackage[T1]{fontenc}
\usepackage[utf8]{inputenc}
\usepackage{microtype}
\usepackage{tikz}
\usetikzlibrary{calc, backgrounds, positioning}

\usepackage{booktabs}       
\usepackage{pgfplots}
\usepackage{subcaption}
\usepackage{amsmath}
\usepackage{enumitem}

\newcommand{\ourssys}{\textsc{OurSystem}}
\newcommand{\twostep}{\textsc{Two-Step}}

\title{QMUL-SDS at SCIVER:\\Step-by-Step Binary Classification for Scientific Claim Verification}

\author{Xia Zeng \\
  Queen Mary University of London \\
  \texttt{x.zeng@qmul.ac.uk} \\\And
  Arkaitz Zubiaga \\
Queen Mary University of London \\
  \texttt{a.zubiaga@qmul.ac.uk} \\}

\begin{document}
\maketitle
\begin{abstract}
Scientific claim verification is a unique challenge that is attracting increasing interest. The SCIVER shared task offers a benchmark scenario to test and compare claim verification approaches by participating teams and consists in three steps: relevant abstract selection, rationale selection and label prediction. In this paper, we present team QMUL-SDS's participation in the shared task. We propose an approach that performs scientific claim verification by doing binary classifications step-by-step. We trained a BioBERT-large classifier to select abstracts based on pairwise relevance assessments for each <claim, title of the abstract> and continued to train it to select rationales out of each retrieved abstract based on <claim, sentence>. We then propose a two-step setting for label prediction, i.e. first predicting ``NOT\_ENOUGH\_INFO'' or ``ENOUGH\_INFO'', then label those marked as ``ENOUGH\_INFO'' as either ``SUPPORT'' or ``CONTRADICT''. Compared to the baseline system, we achieve substantial improvements on the dev set. As a result, our team is the No. 4 team on the leaderboard. 
\end{abstract}

\section{Introduction}

\begin{figure}[h!]
\centering
\begin{tikzpicture}[
roundnode/.style={circle, draw=green!60, fill=green!5, very thick, minimum size=7mm},
squarednodey/.style={rectangle, draw=yellow!60, fill=yellow!5, very thick, minimum size=5mm},
squarednoder/.style={rectangle, draw=red!60, fill=red!5, very thick, minimum size=5mm},
squarednodeg/.style={rectangle, draw=green!60, fill=green!5, very thick, minimum size=5mm},
]
\node[roundnode]        (claim)       {claim $c$};
\node[squarednodey]      (abstracts0)    [below=of claim]{Top $K$ abstracts};
\node[squarednodey]      (abstracts1)    [below=of abstracts0]{Identified abstracts};
\node[squarednodey]      (rationales)    [below=of abstracts1]{Identified rationales};
\node[squarednoder]      (label0)    [below=of rationales]{"Enough\_Info" classification};
\node[squarednodeg]      (NEI)    [right=of label0]{verdict $NEI$};
\node[squarednoder]      (label1)    [below=of label0]{"Support" classification};
\node[squarednodeg]      (S)    [below=of label1]{verdict $S$};
\node[squarednodeg]      (C)    [right=of label1]{verdict $C$};

\draw (claim.south) 
      edge[->] node[]{TF-IDF similarity ranking} 
      (abstracts0.north);
\draw (abstracts0.south) 
      edge[->] node[]{BioBERT abstract classification} 
      (abstracts1.north);
\draw (abstracts1.south) 
      edge[->] node[]{BioBERT rationale classification} 
      (rationales.north);
\draw (rationales.south) 
      edge[-] node[yshift=-1em]{} 
      (label0.north);
\draw (label0.east) 
      edge[->] node[yshift=-1em]{negative} 
      (NEI.west);
\draw (label0.south) 
      edge[->] node[]{positive} 
      (label1.north);
\draw (label1.south) 
      edge[->] node[]{positive} 
      (S.north);
\draw (label1.east) 
      edge[->] node[yshift=-1em]{negative} 
      (C.west);
\begin{pgfonlayer}{background}
\draw [join=round,blue,dotted] ($(abstracts0.north west) + (-2em, 2em)$) rectangle ($(abstracts1.south east) + (2em, 0em)$) node[, xshift=2em, yshift=8em]{\textbf{Abstract Retrieval}};
\draw [join=round,blue,dotted] ($(rationales.north west) + (-2em, 2em)$) rectangle ($(rationales.south east) + (2em, -1em)$) node[, xshift=2em, yshift=4.5em]{\textbf{Rationale Selection}};
\draw [join=round,blue,dotted] ($(label0.north west) + (-1em, 1em)$) rectangle ($(label1.south east) + (2em, -2em)$) node[, xshift=1em, yshift=8.5em]{\textbf{Label Prediction}};
\end{pgfonlayer}
\end{tikzpicture}
\caption{Overview of our step-by-step binary classification system. $NEI$ stands for ``NOT\_ENOUGH\_INFO'', $C$ stands for ``CONTRADICT'' and $S$ stands for ``SUPPORT''. Given claim $c$, our system first retrieves top K TF-IDF similarity abstracts out of the corpus, then uses a BioBERT binary classifier to further identify desired abstracts on top of that. With retrieved abstracts, our system then uses another BioBERT binary classifier to select rationales. We finally do label prediction in a two-step fashion, i.e. first make verdicts on ``ENOUGH\_INFO'' or not and, if positive, then make verdicts on ``SUPPORT'' or not. }
\label{figure:overview}
\end{figure}
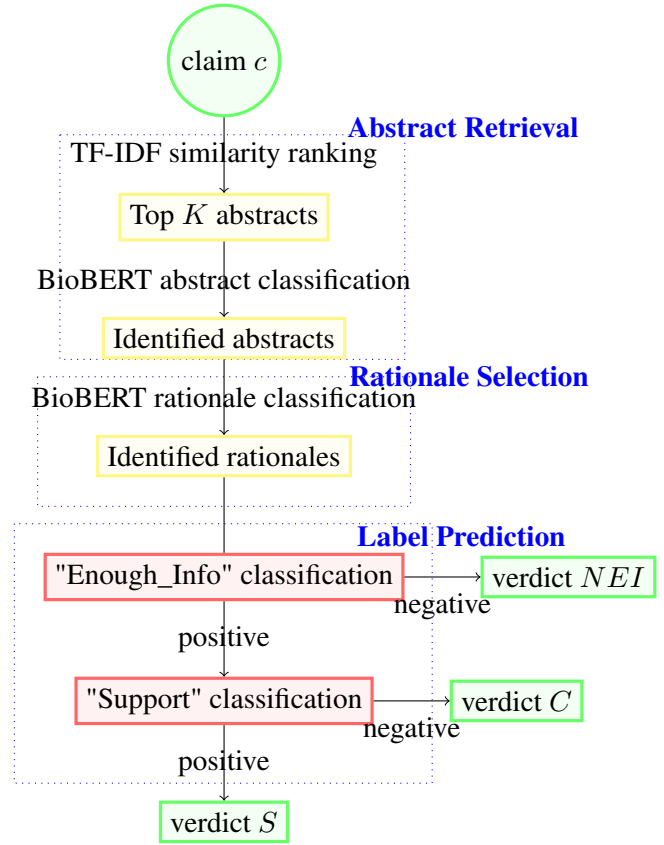

As online content continues to grow at an unprecedented rate, the spread of false information online increases the potential of misleading people and causing  harm. Where the volume of information shared online is difficult to be managed by human fact-checkers, this leads to an increasing demand on automated fact-checking, which is formulated by researchers as ‘the assignment of a truth value to a claim made in a particular context’\cite{vlachos_fact_2014}.

Though a body of research focuses on conducting fact-checking in the politics domain, scientific claim verification has also gained increasing interest in the context of the ongoing COVID-19 pandemic. The SCIVER shared task provides a valuable benchmark to build and evaluate systems performing scientific claim verification. Given a scientific claim and a corpus of over 5000 abstracts, the task consists in (i) identifying abstracts relevant to the claim, (ii) delving into the abstracts to select evidence sentences relevant to the claim, and (iii) subsequently predicting claim veracity.

This paper presents and analyses team QMUL-SDS's participation in the SCIVER shared task. In particular, we explore creative approaches of solving the challenge with limited resources. Figure \ref{figure:overview} provides an overview of our system. While many other systems make use of external datasets, e.g. FEVER \cite{thorne_fever_2018}, our system focuses on efficient use of the SCIFACT dataset \cite{wadden_fact_2020}. Furthermore, in the interest of keeping the efficiency of our system, we limit our model choices to the size of RoBERTa-large \cite{liu_roberta_2019}, ruling out for example GPT-3 \cite{brown_language_2020} and T5 \cite{raffel_exploring_2020}, which were used in other participating systems. More specifically, our system mainly uses RoBERTa \cite{liu_roberta_2019} and BioBERT \cite{lee_biobert_2020}. The latter is pre-trained on biomedical text and therefore is very close to our target domain. With improved pipeline design, our system shows competitive performance with limited computing resources, achieving the 6th position in the task and ranked 4th when distinct teams are considered. \footnote{Code is available \href{https://github.com/XiaZeng0223/sciverbinary.git}{here}.}

\section{Related Work}

Several approaches have been proposed to perform scientific claim verification in the three-step settings proposed in SCIVER.

Upon publication of the SCIFACT dataset \cite{wadden_fact_2020}, the authors introduced VERISCI as a baseline system. It is a pipeline with three modules: abstract retrieval, rationale selection and label prediction. The abstract retrieval module returns the top $K$ highest-ranked abstracts determined by the TF-IDF similarity between each abstract and the claim at hand. The rationale selection module trains a RoBERTa-large model to compute relevance scores with a sigmoid function and then selects sentences whose relevance scores are higher than the threshold $T$. The label prediction module trains a RoBERTa-large model to do three-way classification regarding sentence-pairs, where the candidate labels are "SUPPORT", "CONTRADICT" and "NOT\_ENOUGH\_INFO". Empirically the system set the $K$ value to 3 and the $T$ value to 0.5. Due to its inspiring design, reasonable performance and good efficiency, in this paper we take VERISCI system as our baseline.

After the publication of the SCIFACT dataset, several approaches have been published, some of which chose to participate in the SCIVER shared task. We next discuss the top 3 ranked entries. The VERT5ERINI system \cite{pradeep_scientific_2020} ranked 1st on the leaderboard. This system first retrieves a shortlist of top 20 abstracts by using the BM25 ranking score \cite{robertson_okapi_1994}, which is then fed into a T5 model to rerank and retrieve the top 3 abstracts; it then trains a T5 model to calculate relevance scores for each sentence, on which a threshold of 0.999 is applied to select rationales; it finally trains a T5 model to do three-way classification for predicting labels. This system has demonstrated the performance advantages of using T5, a model that is substantially bigger than other language models. 

The ParagraphJoint system \cite{li_paragraph-level_2021} ranked 2nd on the leaderboard. It first uses BioSentVec \cite{chen_biosentvec_2019} to retrieve the top $K$ abstracts and then jointly trains a RoBERTa-large model to do rationale selection and label prediction in a multi-task learning setting. The system is first trained on the FEVER dataset and then trained on SCIFACT dataset. Its application of multi-task learning techniques proved to be very successful and inspires further research in this direction.

The team who ranked 3rd on the leaderboard, Law \& Econ \cite{stammbach_e-fever_nodate}, fine-tuned their e-FEVER system on SCIFACT dataset, which requires usage of GPT-3 and training on FEVER dataset. Despite the big difference on model sizes, our system achieves close performance to the e-FEVER system on the leaderboard.

\section{Approach}
Following the convention of automated fact-checking systems \cite{thorne_fever_2018} and the VERISCI baseline system, we explore novel ways of tackling the challenge by handling the three subtasks: abstract retrieval, rationale selection and label prediction. 

\subsection{Abstract Retrieval}

Abstract retrieval is the task of retrieving relevant abstracts that can support the prediction of a claim's veracity. Inspired by the baseline system, which retrieves the top $K$ $(K=3)$ abstracts with the highest TF-IDF similarity to the claim, initially we attempted a similar method with a state-of-the-art similarity metric, i.e., BERTscore \cite{zhang_bertscore_2020}. It computes token similarity using BERT-based contextual embeddings. However, the results we achieved were not satisfactory\footnote{See detailed results in appendix \ref{sec:appendix}} and was ruled out in subsequent experiments.

Instead of completely relying on available metrics, we investigated performing abstract retrieval in a supervised manner. In contrast to previous work \cite{pradeep_scientific_2020} which performed reranking, we formulate it as a binary classification problem. We first empirically limit the corpus to the top 30 abstracts with highest TF-IDF similarity to the claim. We fine-tuned a BioBERT model \cite{lee_biobert_2020} with a linear classification head, which we name as the BioBERT classifier thereafter, to do binary classification on the top 30 TF-IDF abstracts, i.e. predicting whether the abstract at hand is correctly identified for the claim at hand given the pairwise input <claim $c$, title $t$ of the abstract>. Due to the input length limits of BERT models, we only use the title of the abstract at this stage, assuming that the title represents a good summary of the abstract.

\subsection{Rationale Selection}

Rationale Selection is the task of selecting rationale sentences out of the retrieved abstracts. To avoid manually tuning the threshold on various settings like the baseline system, we address the problem as a binary classification task in a very similar manner to the last step. We continued training the BioBERT classifier inherited from the abstract retrieval step to do rationale selection, i.e. making binary predictions on whether the sentence at hand is correctly identified for the claim at hand given sentence pair <claim $c$, sentence $s$>. As our classifier model only outputs binary predictions with its linear head on individual sentence pair cases, there is no need to apply various ranking thresholds. Aiming to achieve better overall pipeline performance, our models are trained on abstracts retrieved in the first step, rather than oracle abstracts.

\subsection{Label Prediction}
\label{ssec:label-pred}

Label prediction is the task of predicting the veracity label given the target claim and rationale sentences selected in the preceding step of the pipeline. A good selection of relevant abstracts and rationales therefore is vital in the capacity of the veracity label prediction system. 

The baseline system we initially implemented trained a RoBERTa-large model to do three-way classification into one of ``NOT\_ENOUGH\_INFO'', ``SUPPORT'' and ``CONTRADICT''. We observed that, while the model was in general fairly accurate, it performed poorly in predicting the "CONTRADICT" class due to the scarcity of training data pertaining to this class. However, it is known that claims belonging to the ``CONTRADICT'' class are particularly difficult to collect, and that automated fact-checking datasets tend to create them synthetically by manually mutating naturally occurring claims originally pertaining to the ``SUPPORT'' class \cite{thorne_fever_2018, wadden_fact_2020, sathe_automated_2020}. With the aim of improving model performance on this class without using extra data, we try to decrease wrong predictions accumulated by wrong predictions on the other labels. For instance, the model may predict a claim to be ``NOT\_ENOUGH\_INFO'' while it should be ``CONTRADICT'', which makes it a false positive for the ``NOT\_ENOUGH\_INFO'' class and a true negative for the ``CONTRADICT'' class. If the model has better performance on the ``NOT\_ENOUGH\_INFO'' predictions, it would in turn help the performance on the ``CONTRADICT'' class.

Hence, we explore label prediction within a two-step setting. First, we merge claims from the ``SUPPORT'' and ``CONTRADICT'' classes as ``ENOUGH\_INFO''. With this altered dataset, we train a RoBERTa-large model as a neutral detector to do binary classification into ``ENOUGH\_INFO'' or ``NOT\_ENOUGH\_INFO''. Second, we merge data from ``NOT\_ENOUGH\_INFO'' and ``CONTRADICT'' to be ``NOT\_SUPPORT'' and train another RoBERTa-large model as a support detector to do binary classification on ``SUPPORT'' or ``NOT\_SUPPORT''. Finally, when doing predictions, we first use the neutral detector to predict ``ENOUGH\_INFO'' or ``NOT\_ENOUGH\_INFO'' and only if the first prediction is ``ENOUGH\_INFO'' we use the support detector to predict ``SUPPORT'' or ``NOT\_SUPPORT''. We take ``NOT\_SUPPORT'' instances as equivalent to ``CONTRADICT'' instances in the three-way classification.

\section{Results}
We perform various experiments on the SCIFACT dataset to identify the best models and techniques to be submitted to the task. Unless explicitly specified, models are trained on the SCIFACT's train set and evaluated on the SCIFACT's dev set.

\subsection{Abstract Retrieval}

We limit the candidate abstracts to the top 30 with the highest TF-IDF similarity scores, as this setting achieves a high recall of 91.39\%. With our binary classification method, we experimented with BioBERT models that are pre-trained on close domain texts \cite{lee_biobert_2020}. To explore the potentials of adapting pre-trained language models to the current settings, we also conducted task adaptive pre-training \cite{gururangan_dont_2020} on the SCIFACT corpus with BioBERT-base for 50 epochs with batch size 1, which leads to a final perplexity of 2.68. This parameter choice is made primarily based on our limited time and computational resources for the SCIVER shared task participation. Further extensive exploration may lead to interesting results. This model is denoted as BioBERT-base*.

Table \ref{table:abstract} reports performance of the baseline, BioBERT-base, BioBERT-base* and BioBERT-large models on abstract retrieval. The baseline directly retrieves the top 3 abstracts with highest TF-IDF similarity, which is also the method used in the VERISCI system \cite{wadden_fact_2020}. We also report abstract level pipeline performance with baseline rationale selector and baseline label predictor to demonstrate its substantial impact on pipeline performance.

Our method achieves noticeable improvements over the baseline by largely decreasing the false positive rate. More specifically, BioBERT-base has the highest precision score, BioBERT-base* has highest F1 score and BioBERT-large has the highest recall score. With increased model size, BioBERT-large has gained significant improvements on recall but suffers with a precision drop compared to BioBERT-base and BioBERT-base*, which may suggest model underfitting. Overall our approach leads to an approximate 10\% increase over the baseline approach on abstract level downstream performance. 

\begin{table}[t]
\centering\centering{}{
\small
\begin{tabular}{l|lll}
\toprule
\multicolumn{4}{c}{Abstract Retrieval}\\
\midrule
Method & P & R & F1\\
\toprule
Baseline & 16.22 & 69.86 & 26.33 \\
BioBERT-base & \textbf{83.23} & 64.11 & 72.43 \\
BioBERT-base* & 81.61 & 67.94 & \textbf{74.15} \\
BioBERT-large & 62.75 & \textbf{74.16} & 67.98 \\
\bottomrule
\toprule
\multicolumn{4}{c}{Downstream Performance} \\
\midrule
\multicolumn{4}{c}{Abstract Level Label Only} \\
\midrule
Method & P & R & F1\\
\toprule
Baseline & 56.42 & 48.33 & 52.06 \\
BioBERT-base & 84.30 & 48.80 & 61.82 \\
BioBERT-base* & \textbf{84.92} & 51.20 & \textbf{63.88} \\
BioBERT-large & 79.71 & \textbf{52.63} & 63.40 \\
\bottomrule
\midrule
\multicolumn{4}{c}{Abstract Level Label + Rationale} \\
\midrule
Method & P & R & F1\\
\toprule
Baseline & 54.19 & 46.41 & 50.00 \\
BioBERT-base & 81.82 & 47.37 & 60.00 \\
BioBERT-base* & \textbf{82.54} & 49.76 & \textbf{62.09} \\
BioBERT-large & 76.81 & \textbf{50.72} & 61.10 \\
\bottomrule
\end{tabular}
}
\vspace{-0.1cm}
\caption{Comparison of abstract retrieval methods on the dev set of SCIFACT.}
\label{table:abstract}
\vspace{-0.25cm}
\end{table}

\subsection{Rationale Selection}

In order to improve the overall design of the system, we trained our rationale selection models with abstracts retrieved by our abstract retrieval module rather than oracle abstracts. We use abstracts retrieved by BioBERT-large due to its highest recall score. In this step, we experiment with our binary classification approach to identify rationale sentences from retrieved abstracts for the claim at hand. Given a sentence-pair <claim $c$, sentence $s$>, the model, which was trained to do abstract selection in last step, is now trained to predict whether the sentence at hand is correctly identified for the claim at hand.

Table \ref{table:rationale} reports results of the baseline, BioBERT-base, BioBERT-base* and BioBERT-large models on rationale selection. We also present sentence level pipeline performance with oracle cited abstracts \footnote{It includes abstracts that are of "SUPPORT", "CONTRADICT" and "NOT\_ENOUGH\_INFO" relations to the claims' veracity. It is also referred as oracle abstracts with NOT\_ENOUGH\_INFO (NEI) setting in SCIFACT dataset paper.} and baseline label predictor.

Our method leads to an increase in precision score, a small decrease in recall score and a small increase in F1 score. Interestingly, the three BioBERT variants don't show clear performance differences, despite substantial differences in model sizes. A small improvement on downstream sentence-level performance is achieved overall.

\begin{table}[t]
\centering\centering{}{
\small
\begin{tabular}{l|lll}
\toprule
\multicolumn{4}{c}{Sentence Selection}\\
\midrule
Method & P & R & F1\\
\toprule
Baseline & 64.99 & \textbf{70.49} & 67.63 \\
BioBERT-base & \textbf{77.97} & 62.84 & \textbf{69.59} \\
BioBERT-base* & 74.38 & 65.03 & 69.39 \\
BioBERT-large & 77.08 & 63.39 & 69.57 \\
\bottomrule
\toprule
\multicolumn{4}{c}{Downstream Performance} \\
\midrule
\multicolumn{4}{c}{Sentence Level Selection Only} \\
\midrule
Method & P & R & F1\\
\toprule
Baseline & 74.48 & \textbf{59.02} & 65.85 \\
BioBERT-base & \textbf{83.81} & 56.56 & 67.54 \\
BioBERT-base* & 80.84 & 57.65 & 67.30 \\
BioBERT-large & 80.75 & 58.47 & \textbf{67.83} \\
\bottomrule
\midrule
\multicolumn{4}{c}{Sentence Level Selection + Label} \\
\midrule
Method & P & R & F1\\
\toprule
Baseline & 66.90 & 53.01 & 59.15 \\
BioBERT-base & \textbf{74.90} & 50.55 & 60.36 \\
BioBERT-base* & 72.41 & 51.64 & 60.29 \\
BioBERT-large & 72.08 & \textbf{52.19} & \textbf{60.54} \\
\bottomrule
\end{tabular}
}
\vspace{-0.1cm}
\caption{Comparison of rationale selection methods on the dev set of SCIFACT.}
\label{table:rationale}
\vspace{-0.25cm}
\end{table}

\subsection{Label Prediction}

For label prediction, we use the two-step approach that leverages RoBERTa-large as described in \S \ref{ssec:label-pred}. This approach is denoted as \textsc{Two-Step} thereafter. Table \ref{table:label1} reports performance results for the label prediction task with oracle cited abstracts and oracle rationales. The baseline is the RoBERTa-large three-way classifier used on VERISCI. Our \textsc{Two-Step} method leads to a 4\% increase in accuracy, macro-F1 and weighted-F1 over the baseline. We further present confusion matrices for each system for analysis, where C stands for ``CONTRADICT'', N stands for ``NOT\_ENOUGH\_INFO'' and S stands for ``SUPPORT''. As the confusion matrix shows, our method successfully improves the overall predictions on the ``CONTRADICT'' class without leveraging extra data.

Furthermore, Table \ref{table:label2} reports results on the abstract-level label prediction with various settings of upstream modules. Interestingly, both methods show noticeably decreased performance when given an evidence of lower quality. From the oracle evidence to the evidence retrieved by our system, the baseline module's F1 performance dropped by 19.70\% and the \textsc{Two-Step} module dropped by 20.26\% in absolute values; from the oracle evidence to the evidence retrieved by the baseline system, the baseline module's F1 score dropped by 30.14\% and the \textsc{Two-Step} module dropped by 37.26\% in absolute values.

Despite that, our \textsc{Two-Step} method always outperforms the baseline method when given improved evidence. Its F1 score is 2.02\% - 2.58\% higher than the baseline on improved evidence retrieval settings. When given oracle cited abstracts and oracle rationales, our method achieves 84.78\% F1 score. 

\begin{table}[t]
\centering\centering{}{
\small
\begin{tabular}{c|ccc}
\toprule
\multicolumn{4}{c}{Label Prediction Performance}\\
\midrule
Method & Accuracy & Macro-F1 & Weighted-F1 \\
\toprule
Baseline & 81.93 & 80.19 & 81.85 \\
\twostep & \textbf{85.98} & \textbf{84.69} & \textbf{85.84} \\
\bottomrule
\toprule
\multicolumn{4}{c}{Confusion Matrix of Baseline} \\
\midrule
  & C & N & S\\
\toprule
C & 47 & 17 & 7 \\
N & 6 & 104 & 2 \\
S & 8 & 18 & 112 \\
\bottomrule
\toprule
\multicolumn{4}{c}{Confusion Matrix of \twostep} \\
\midrule
 & C & N & S\\
\toprule
C & 53 & 7 & 11 \\
N & 2 & 107 & 3 \\
S & 12 & 10 & 116 \\
\bottomrule
\end{tabular}
}
\vspace{-0.1cm}
\caption{Comparison of label prediction methods with oracle cited abstracts and oracle rationales.}
\label{table:label1}
\vspace{-0.25cm}
\end{table}

\begin{table}[t]
\centering\centering{}{
\small
\begin{tabular}{l|lll}
\toprule
\multicolumn{4}{c}{Oracle Abstract + Oracle Rationale} \\
\midrule
Method & P & R & F1\\
\toprule
Baseline & \textbf{90.75} & 75.12 & 82.20 \\
\twostep & {88.54} & \textbf{81.33} & \textbf{84.78} \\
\bottomrule
\toprule
\multicolumn{4}{c}{OurSystem Abstract + OurSystem Rationale} \\
\midrule
Method & P & R & F1\\
\toprule
Baseline & \textbf{76.92} & 52.63 & 62.50 \\
\twostep & {73.62} & \textbf{57.42} & \textbf{64.52} \\
\bottomrule
\toprule
\multicolumn{4}{c}{Baseline Abstract + Baseline Rationale}\\
\midrule
Method & P & R & F1\\
\toprule
Baseline & \textbf{56.42} & 48.32 & \textbf{52.06} \\
\twostep & 43.31 & \textbf{52.63} & 47.52 \\
\bottomrule
\end{tabular}
}
\vspace{-0.1cm}
\caption{Comparison of label prediction methods with various upstream modules.}
\label{table:label2}
\vspace{-0.25cm}
\end{table}

\subsection{Full Pipeline}

Table \ref{table:pipeline1} reports full pipeline performance on the SCIFACT dev set. The baseline is the VERISCI system. We compare pipeline systems with different evidence retrieval models, i.e., BioBERT-base, BioBERT-base* and BioBERT-large, combined with the two-step label predictor using RoBERTa-large.

Overall our system achieves substantial improvements over the baseline. Across the evaluation metrics, our precision scores are 15.75\%-23.37\% higher than the baseline system, recall scores are 3.82\%-14.21\% higher and F1 scores are 10.11\%-16.08\% higher than the baseline in terms of absolute values. Interestingly, BioBERT-base obtains the highest precision score, BioBERT-base* the highest recall score and BioBERT-large the highest F1 for most of metrics.  

\begin{table}[t]
\centering\centering{}{
\small
\begin{tabular}{l|lll}
\toprule
\multicolumn{4}{c}{Label Only} \\
\midrule
System & P & R & F1\\
\toprule
Baseline & 56.42 & 48.33 & 52.06 \\
BioBERT-base + \twostep & \textbf{79.56} & 52.15 & 63.00 \\
BioBERT-base* + \twostep & 78.91 & 55.50 & \textbf{65.17} \\
BioBERT-large + \twostep & 73.62 & \textbf{57.42} & 64.52 \\
\bottomrule
\toprule
\multicolumn{4}{c}{Label+Rationale}\\
\midrule
System & P & R & F1\\
\toprule
Baseline & 54.19 & 46.41 & 50.00 \\
BioBERT-base + \twostep & \textbf{75.91} & 49.76 & 60.11 \\
BioBERT-base* + \twostep & 73.47 & 51.67 & 60.67 \\
BioBERT-large + \twostep & 69.94 & \textbf{54.55} & \textbf{61.29} \\
\bottomrule
\toprule
\multicolumn{4}{c}{Selection Only} \\
\midrule
System & P & R & F1\\
\toprule
Baseline & 54.27 & 43.44 & 48.25 \\
BioBERT-base + \twostep & \textbf{77.64} & 52.19 & 62.42 \\
BioBERT-base* + \twostep & 72.00 & 54.10 & 61.78 \\
BioBERT-large + \twostep & 72.76 & \textbf{57.65} & \textbf{64.33} \\
\bottomrule
\toprule
\multicolumn{4}{c}{Selection+Label}\\
\midrule
System & P & R & F1\\
\toprule
Baseline & 48.46 & 38.80 & 43.10 \\
BioBERT-base + \twostep & \textbf{68.29} & 45.90 & 54.90 \\
BioBERT-base* + \twostep & 64.00 & 48.09 & 54.92 \\
BioBERT-large + \twostep & 64.83 & \textbf{51.37} & \textbf{57.32}\\
\bottomrule
\end{tabular}
}
\vspace{-0.1cm}
\caption{Comparison of full pipeline performance on the dev set of SCIFACT.}
\label{table:pipeline1}
\vspace{-0.25cm}
\end{table}

Table \ref{table:pipeline2} compares full pipeline performance on SCIFACT test set with models trained on the combination of SCIFACT train set and dev set. We used BioBERT-large evidence selector and two-step label predictor as our system due to its overall best performance. This submission ranked No. 6 on the leaderboard.

\begin{table}[t]
\centering\centering{}{
\small
\begin{tabular}{l|lll}
\toprule
\multicolumn{4}{c}{Label Only} \\
\midrule
System & P & R & F1\\
\toprule
Baseline & 47.51 & 47.30 & 47.40\\
\ourssys & \textbf{74.32} & \textbf{49.55} & \textbf{59.46} \\
\bottomrule
\toprule
 \multicolumn{4}{c}{Label+Rationale}\\
\midrule
System & P & R & F1\\
\toprule
Baseline & 46.61 & 46.40 & 46.50\\
\ourssys & \textbf{72.97} & \textbf{48.65} & \textbf{58.38} \\
\bottomrule
\toprule
\multicolumn{4}{c}{Selection Only} \\
\midrule
System & P & R & F1\\
\toprule
Baseline & 44.99 & 47.30 & 46.11\\
\ourssys & \textbf{81.58} & \textbf{58.65} & \textbf{68.24} \\
\bottomrule
\toprule
\multicolumn{4}{c}{Selection+Label}\\
\midrule
System & P & R & F1\\
\toprule
\hspace{-4.5pt} Baseline & 38.56 & 40.54 & 39.53 \\
\hspace{-4.5pt} \ourssys & \textbf{66.17} & \textbf{47.57} & \textbf{55.35}\\
\bottomrule
\end{tabular}
}
\vspace{-0.1cm}
\caption{Full pipeline performance on SCIFACT's test set. \ourssys uses BioBERT-large for abstract retrieval and rationale selection with two-step label prediction, all trained on trained set and dev set.}
\label{table:pipeline2}
\vspace{-0.25cm}
\end{table}

\section{Discussion and Future Work}

Our intuitive step-by-step binary classification system achieves substantial improvements over the baseline without demanding additional data or extra large models. 

An improved evidence retrieval module has made the main contributions to the performance boost. Our system makes an effort to improve the abstract retrieval module after applying a scalable traditional information retrieval weighting scheme, TF-IDF. Instead of handling it as a re-ranking task and manually selecting thresholds \cite{pradeep_scientific_2020}, we formulate it as a binary classification task, which makes better use of the available training data and decreases the false positive rate effectively. When applying a similar approach to rationale selection, our model, which is only trained on the SCIFACT dataset, still achieves improvements over the baseline model, which makes use of the FEVER dataset first. Furthermore, our model is less dependent on parameters than other systems, which is ideal in practical settings where one would like to apply the model on new datasets without having to find the best parameters for the dataset at hand.

In addition, our \textsc{Two-Step} label prediction module also makes positive contributions to overall improvements. The difference on the label prediction performance is very noticeable on different upstream settings. Unsurprisingly, both methods have the best performance with F1 scores higher than 80\% on the oracle setting, which is the closest to their training data. Interestingly, this performance fluctuation leads to the following observation: a label prediction module that has better performance on the oracle evidence doesn't necessarily have better performance when given the incorrect evidence. Regarding our \textsc{Two-Step} label prediction method, it shows that our neutral detector is not robust enough on the pipeline setting. One possible solution is to train it on evidence retrieved by previous modules rather than on the oracle evidence so that it learns to optimise for the pipeline setting.

Nevertheless, this problem is inevitable for a pipeline system that has multiple machine learning modules, as errors in each of the modules will accumulate throughout the pipeline. A better system design is desired such that it tackles the challenge in a more systematic way. A promising approach is to train a model to learn three subtasks in a multitask learning manner so that it may optimise for better overall performance.

\section{Conclusions}
In this paper, we proposed a novel step-by-step binary classification approach for the SCIVER shared task. Our submission achieved an F1 score of 55.35\% on the test set, ranking 6th among all the submissions and 4th among all the teams. We show that (1) concerning evidence retrieval, a classification based approach is better than a ranking based approach with manual thresholds; (2) two-step binary label prediction has better performance than three-way label prediction with limited training data; (3) a more systematic design of automated fact-checking system is desired.

\section*{Acknowledgements}
This work was supported by the Engineering and Physical Sciences Research Council (grant  EP/V048597/1). Xia Zeng is funded by China Scholarship Council (CSC). This research utilised Queen Mary’s Apocrita HPC facility, supported by QMUL Research-IT. \url{http://doi.org/10.5281/zenodo.438045}

\bibliography{reference}
\bibliographystyle{acl_natbib}

\appendix

\section{Appendix}
\label{sec:appendix}

Table \ref{table:BERTscore} reports performance of using BERTscore as a metric to do abstract retrieval. We chose DistilBERT as the BERT model for global ranking for efficiency reasons, which was ran on a simgle GPU for approximately 36 hours and it turned out to be worse than TF-IDF. 

We then tried various relevant BERT variants to do reranking out of the top 30 abstracts with the highest TF-IDF similarity. In general, with reasonable large models that are trained on relevant tasks, results are better than TOP 3 TF-IDF. However, the improvements remain trivial and it is not comparable to our classification approach.

\begin{table}[h]
\centering\centering{}{
\small
\begin{tabular}{l|lll}
\toprule
\multicolumn{4}{c}{TOP K Global Ranking with DistilBERT}\\
\midrule
Method & P & R & F1\\
\toprule
TF-IDF TOP 1 & 60.11 & 54.07 & 56.93 \\
BERTscore TOP 1 & 51.06 & 45.93 & 48.36 \\
TF-IDF TOP 3 & 25.89 & 69.86 & 37.78 \\
BERTscore TOP 3 & 23.58 & 63.64 & 34.41 \\
TF-IDF TOP 30 & 03.39 & 91.39 & 06.54 \\
BERTscore TOP 30 & 03.26 & 88.04 & 06.29 \\
\bottomrule
\toprule
\multicolumn{4}{c}{TOP 3 BERTscore Reranking under TOP 30 TF-IDF} \\
\midrule
Model & P & R & F1\\
\toprule
BERT-tiny & 23.94 & 64.59 & 34.93 \\
SciBERT & 25.89 & 69.86 & 37.78 \\
BioBERT-base & 28.37 & 76.56 & 41.40 \\
BioBERT-large & 26.60 & 71.77 & 38.81 \\
RoBERTa-large rationale selector & 20.39 & 55.02 & 29.75 \\
RoBERTa-large label predictor  & 25.89 & 69.86 & 37.78 \\
\bottomrule
\end{tabular}
}
\vspace{-0.1cm}
\caption{BERTscore abstract retrieval performance on the dev set of SCIFACT.}
\label{table:BERTscore}
\vspace{-0.25cm}
\end{table}

\end{document}